\documentclass{article}
\usepackage{iclr2026_conference}
\usepackage{times}
\usepackage{hyperref}
\usepackage{url}
\usepackage{graphicx}
\usepackage{amsmath,amssymb}
\usepackage{booktabs}
\usepackage{multirow}
\usepackage{xcolor}
\usepackage{listings}
\usepackage{algorithm}
\usepackage{algpseudocode}
\usepackage{enumitem}
\usepackage[T1]{fontenc}
\graphicspath{{./figs}}

\title{Flow-Factory: A Unified Framework for \\ Reinforcement Learning in Flow-Matching Models}

\author{%
\textbf{Bowen Ping}\textsuperscript{1},
\textbf{Chengyou Jia}\textsuperscript{1},
\textbf{Minnan Luo}\textsuperscript{1},
\textbf{Hangwei Qian}\textsuperscript{2},
\textbf{Ivor Tsang}\textsuperscript{2}\\[3pt]
\textsuperscript{1}Xi'an Jiaotong University \quad
\textsuperscript{2}CFAR and IHPC, A*STAR\\[3pt]
\texttt{jayceping6@gmail.com}
}

\iclrfinalcopy % Uncomment for camera-ready version, but NOT for submission.
\begin{document}

\maketitle
% \lhead{Preprint}
\begin{abstract}
Reinforcement learning has emerged as a promising paradigm for aligning diffusion and flow-matching models with human preferences,
yet practitioners face fragmented codebases, model-specific implementations, and engineering complexity.
We introduce \textbf{Flow-Factory},
a unified framework that decouples algorithms, models, and rewards through through a modular, registry-based architecture.
This design enables seamless integration of new algorithms and architectures, as demonstrated by our support for GRPO, DiffusionNFT, and AWM across Flux, Qwen-Image, and WAN video models.
By minimizing implementation overhead, Flow-Factory empowers researchers to rapidly prototype and scale future innovations with ease.
Flow-Factory provides production-ready memory optimization, flexible multi-reward training, and seamless distributed training support.
The codebase is available at \url{https://github.com/X-GenGroup/Flow-Factory}.
\end{abstract}

\section{Introduction}
\label{sec:intro}

Recent advances in diffusion and flow-matching models have revolutionized generative modeling across images, videos, and multimodal content~\citep{song2020score,lipman2022flow,liu2022flow}.
These models have demonstrated remarkable sample quality and training efficiency by learning continuous-time velocity fields that transport noise distributions to data distributions.
However, aligning these powerful generative models with human preferences and real-world constraints remains a fundamental challenge.

Reinforcement learning (RL) has emerged as a promising paradigm for post-training alignment,
enabling models to optimize arbitrary reward signals beyond maximum likelihood training~\citep{black2024ddpo,liu2025flow,xue2025dancegrpo}.
Recent works such as Flow-GRPO~\citep{liu2025flow} and DanceGRPO~\citep{xue2025dancegrpo} have demonstrated the effectiveness of policy gradient methods for flow-matching models by introducing stochastic differential equation (SDE) formulations that enable RL exploration.
Subsequent research has further improved training efficiency through mixed ODE-SDE sampling~\citep{li2025mixgrpo,he2025tempflow},
addressed reward hacking via loss reweighting~\citep{wang2025grpo},
and explored alternative objectives that align RL with diffusion pretraining~\citep{zheng2025diffusionnft,xue2025advantage}.

Despite this rapid algorithmic progress, practitioners face significant barriers when applying RL to flow-matching models:

\paragraph{Fragmented and Coupled Codebases.}
As RL algorithms proliferate with their own SDE formulations and training procedures,
implementations have become increasingly fragmented across isolated repositories.
Moreover, existing codebases frequently couple algorithmic logic with model-specific implementations,
making it difficult to transfer algorithms across architectures without extensive rewriting,
or to evaluate multiple algorithms on the same model under controlled conditions.

\paragraph{Training Inefficiency and Memory Bottlenecks.}
RL fine-tuning of flow-matching models requires repeated forward passes for trajectory sampling and reward computation, leading to significant training overhead.
Additionally, large-scale models with multiple frozen components (text encoders, VAEs) consume substantial GPU memory even when only the transformer backbone is trainable,
limiting batch sizes and hindering scalability on commodity hardware.

\paragraph{Limited Reward Model Flexibility.}
Current frameworks typically support only pointwise reward models that score individual samples independently.
However, emerging algorithms increasingly rely on groupwise rewards that evaluate relative rankings within sample groups~\citep{wang2025pref},
or require combining multiple rewards with different aggregation strategies~\citep{liu2026gdpo}.
The lack of unified reward interfaces restricts algorithmic exploration and multi-objective optimization.

We present \textbf{Flow-Factory}, a unified and extensible framework addressing these challenges through principled software design. Our key contributions include:

\begin{itemize}[leftmargin=*]
\item \textbf{Registry-based modular architecture.} 
Flow-Factory decouples models, algorithms, rewards, and schedulers via a unified plug-and-play registry.
This modular design minimizes implementation overhead,
allowing researchers to integrate new architectures or algorithms with minimal code changes.
Any supported component can be cross-combined through configuration alone,
facilitating principled benchmarking and rapid prototyping.

\item \textbf{Preprocessing-based memory optimization.}
Flow-Factory preprocesses and caches condition embeddings before training, enabling frozen components (text encoders, VAEs) to be offloaded from GPU memory.
During training, only the target transformer resides on GPU, dramatically reducing memory usage while improving throughput by eliminating redundant encoding operations.

\item \textbf{Flexible multi-reward system.}
Flow-Factory provides unified interfaces for both \emph{pointwise} and \emph{groupwise} reward models, with automatic model deduplication and configurable advantage aggregation strategies.
This flexibility supports diverse algorithmic requirements and enables principled multi-reward training with weighted combinations.
\end{itemize}

%% ============================================================
\section{Framework Design}
\label{sec:design}
%% ============================================================

\begin{figure}[t]
    \centering
    \includegraphics[width=\textwidth]{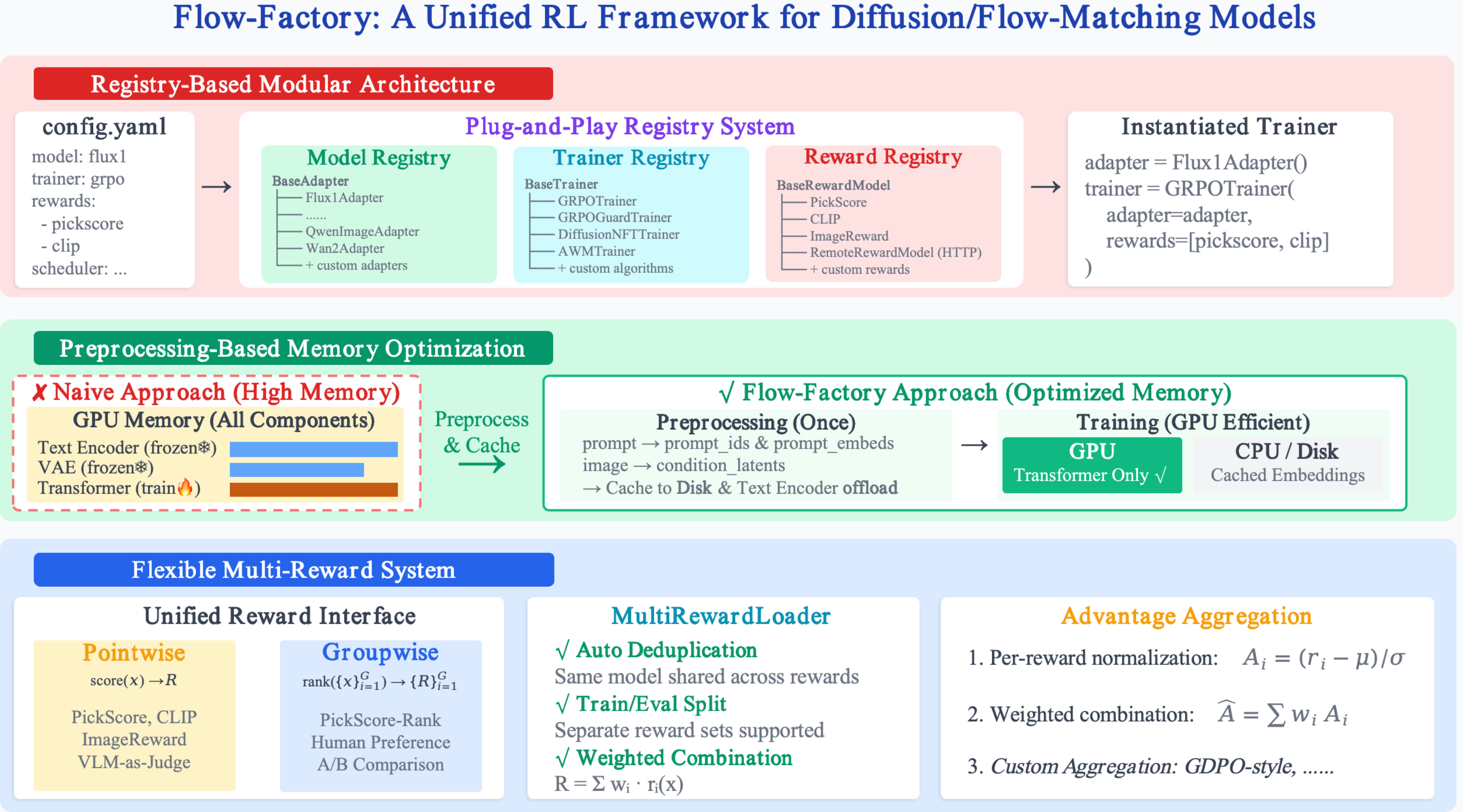}
    \caption{Flow-Factory architecture overview. 
    \textbf{Top:} The registry-based design decouples Models, Trainers, and Rewards, enabling flexible combinations through YAML configuration.
    \textbf{Middle:} Preprocessing optimizes memory by caching embeddings and offloading frozen components.
    \textbf{Bottom:} The multi-reward system supports both pointwise and groupwise rewards with automatic deduplication and configurable advantage aggregation.}
    \label{fig:architecture}
\end{figure}

Figure~\ref{fig:architecture} illustrates the overall architecture of Flow-Factory, which consists of three core design principles: registry-based component decoupling, preprocessing-based memory optimization, and a flexible multi-reward system.

\subsection{Registry-Based Component Decoupling}

Existing RL fine-tuning codebases typically interleave model-specific logic with algorithmic implementations,
forcing researchers to reimplement algorithms when switching architectures.
To address this, Flow-Factory introduces a registry-based architecture with four decoupled component types:
\texttt{BaseAdapter} for model operations (encoding, forward passes, checkpointing),
\texttt{BaseTrainer} for algorithm logic (sampling, advantage computation, optimization),
\texttt{BaseRewardModel} for reward computation,
and \texttt{SDESchedulerMixin} for stochastic sampling with log-probability computation.

By adhering to these standardized interfaces,
integrating a new model or algorithm merely requires implementing the specific abstract methods of the corresponding base class.
This design reduces the integration complexity from $O(M \times N)$ to $O(M + N)$ for $M$ models and $N$ algorithms.
Consequently, components are registered via a global system and instantiated through YAML configurations (Figure~\ref{fig:architecture}, top),
enabling seamless cross-architectural inheritance and allowing practitioners to extend the framework to future innovations with negligible engineering overhead.
This modularity ensures that Flow-Factory remains future-proof as new foundation models and reinforcement learning paradigms emerge.

\subsection{Preprocessing-Based Memory Optimization}

Modern flow-matching pipelines contain large frozen components,
such as text encoders, image encoders, and VAEs,
which consume substantial GPU memory even when only the transformer is trainable,
while RL fine-tuning repeatedly encodes the same prompts across iterations.
Flow-Factory addresses both issues through a two-phase approach:
before training, we precompute and cache all condition embeddings (prompt embeddings, pooled embeddings, VAE latents) to disk;
during training, these cached embeddings are loaded directly, allowing frozen components to be completely offloaded.
As shown in Figure~\ref{fig:architecture} (middle),
this transforms the memory layout to \emph{transformer-only on GPU,},
enabling larger batch sizes while eliminating redundant encoding operations to improve throughput.

\subsection{Multi-Reward System}

Existing frameworks typically support only pointwise rewards that score samples independently,
yet emerging algorithms such as Pref-GRPO~\citep{wang2025pref} require groupwise rewards for ranking-based evaluation
or multi-objective optimization combining heterogeneous signals, which often results redundant model loading when configurations share the same backbone.
Flow-Factory provides a unified reward system (Figure~\ref{fig:architecture}, bottom) with three mechanisms:
(1) \emph{unified interfaces} for both \texttt{PointwiseRewardModel} ($\text{score}(x) \rightarrow \mathbb{R}$) and \texttt{GroupwiseRewardModel} ($\text{rank}(\{x_i\}) \rightarrow \mathbb{R}^k$);
(2) \emph{automatic deduplication} via \texttt{MultiRewardLoader} that loads each unique model once regardless of how many reward configurations reference it;
(3) \emph{configurable advantage aggregation} supporting both weighted-sum and GDPO-style~\citep{liu2026gdpo} per-reward normalization. Implementing new aggregation strategies only requires a new \texttt{compute\_advantages} method in the trainer class.
This design supports diverse algorithmic requirements while minimizing memory overhead and simplifying custom reward integration.

%% ============================================================
\section{Supported Algorithms}
\label{sec:algorithms}
%% ============================================================

Flow-Factory provides unified implementations of state-of-the-art RL algorithms for flow-matching models.
All algorithms share the same model adapter and reward interfaces, enabling direct comparison under controlled conditions.

\subsection{Flow-GRPO and Variants}

Flow-GRPO~\citep{liu2025flow} extends the deterministic ODE sampling of flow-matching models with stochastic noise injection to enable RL exploration.
The standard ODE update $x_{t+dt} = x_t + v_\theta(x_t, t) dt$ is replaced by an SDE formulation:
\begin{equation}
\boldsymbol{x}_{t+dt} = \boldsymbol{x}_t + \left[\boldsymbol{v}_\theta(\boldsymbol{x}_t, t) + \frac{\sigma_t^2}{2t}(\boldsymbol{x}_t + (1-t)\boldsymbol{v}_\theta(\boldsymbol{x}_t, t))\right]dt + \sigma_t\sqrt{dt}\,\epsilon
\label{eq:sde}
\end{equation}
where $\epsilon \sim \mathcal{N}(0, I)$ and $\sigma_t$ controls noise injection.
This formulation enables log-probability computation required for policy gradient optimization.

Flow-Factory implements multiple SDE dynamics through a unified \texttt{SDESchedulerMixin} interface, as summarized in Table~\ref{tab:dynamics}.
Users can switch between formulations via a single configuration parameter, facilitating systematic comparison of their effects on training stability and sample quality.

\begin{table}[htbp]
\centering
\caption{Supported SDE dynamics in Flow-Factory.}
\label{tab:dynamics}
\begin{tabular}{@{}lll@{}}
\toprule
\textbf{Dynamics} & \textbf{Noise Schedule $\sigma_t$} & \textbf{Reference} \\
\midrule
Flow-SDE & $\eta\sqrt{t/(1-t)}$ & Flow-GRPO~\citep{liu2025flow} \\
Dance-SDE & $\eta$ (constant) & DanceGRPO~\citep{xue2025dancegrpo} \\
CPS & $\sigma_{t-1}\sin(\eta\pi/2)$ & FlowCPS~\citep{wang2025coefficients} \\
ODE & $0$ (deterministic) & For NFT~\citep{zheng2025diffusionnft}/AWM~\citep{xue2025advantage} \\
\bottomrule
\end{tabular}
\end{table}

Beyond vanilla GRPO, Flow-Factory supports two important variants.
\textbf{MixGRPO}~\citep{li2025mixgrpo}(i.e., \emph{Flow-GRPO-Fast}) applies SDE updates to only 1--2 selected timesteps while using ODE for the rest, significantly reducing computational cost while maintaining performance.
\textbf{GRPO-Guard}~\citep{wang2025grpo} addresses the negatively-biased ratio distribution inherent in SDE formulations through timestep-dependent loss reweighting, mitigating reward hacking and stabilizing training.

\subsection{DiffusionNFT and AWM}

While GRPO couples sampling dynamics with training timesteps, alternative paradigms exist that decouple them entirely.

\textbf{DiffusionNFT}~\citep{zheng2025diffusionnft} optimizes a contrastive objective directly on the forward flow-matching process:

\begin{equation}
\mathcal{L}_{\text{NFT}} = \mathbb{E}_{\boldsymbol{c}, t} \left[ r \lvert\lvert \boldsymbol{v}_\theta^+(\boldsymbol{x}_t, \boldsymbol{c}, t) - \boldsymbol{v}\rvert\rvert_2^2 + (1 - r) \lvert\lvert \boldsymbol{v}_\theta^-(\boldsymbol{x}_t, \boldsymbol{c}, t) - \boldsymbol{v}\rvert\rvert_2^2 \right]
\label{eq:nft}
\end{equation}
where $\boldsymbol{v}_\theta^+$ and $\boldsymbol{v}_\theta^-$ represent implicit positive and negative policies weighted by a reward $r \in [0, 1]$.
By contrasting these implicit distributions, DiffusionNFT identifies a policy improvement direction without the need for tractable likelihood estimation or SDE-based sampling.

\textbf{Advantage Weighted Matching (AWM)}~\citep{xue2025advantage}
further aligns RL optimization with the flow-matching pretraining objective by weighting the standard velocity matching loss with per-sample advantages:
\begin{equation}
\mathcal{L}_{\text{AWM}} = \mathbb{E}_{t, \epsilon}\left[A(\boldsymbol{x}_0) \cdot \|\boldsymbol{v}_\theta(\boldsymbol{x}_t, t) - (\epsilon - \boldsymbol{x}_0)\|^2\right]
\label{eq:awm}
\end{equation}
where $A(\boldsymbol{x}_0)$ denotes the advantage computed from reward signals.
This formulation incorporates reward-based guidance directly into the velocity matching loss,
effectively aligning the optimization target with the original flow-matching objective.

Since both algorithms decouple training from sampling dynamics, they are inherently solver-agnostic:
any ODE solver can be used for trajectory generation without modifying the training procedure.
Flow-Factory exploits this property by providing configurable timestep sampling strategies (uniform, logit-normal, or discrete) and support for high-order ODE solvers,
enabling practitioners to balance computation cost against sample quality during data collection.

%% ============================================================
\section{Experiments}
\label{sec:experiments}
%% ============================================================

% 目的：验证框架的正确性和效率

\subsection{Experimental Setup}

We evaluate Flow-Factory using FLUX.1-dev~\citep{blackforestlabs2025flux1} with three algorithms: Flow-GRPO~\citep{liu2025flow}, DiffusionNFT~\citep{zheng2025diffusionnft}, and AWM~\citep{xue2025advantage}.
The reward models are chosen as Text-Rendering and PickScore~\citep{kirstain2023pick}, following prior works.

\subsection{Reproduction of Published Results}
To verify the correctness of our implementation and the effectiveness of the decoupled architecture,
we replicate the training processes of three state-of-the-art RL algorithms including Flow-GRPO~\citep{liu2025flow}, DiffusionNFT~\citep{zheng2025diffusionnft}, and AWM~\citep{xue2025advantage} using the FLUX.1-dev model as the common backbone.

\begin{figure}[htbp]
\centering
\begin{minipage}{0.32\textwidth}
    \includegraphics[width=\linewidth]{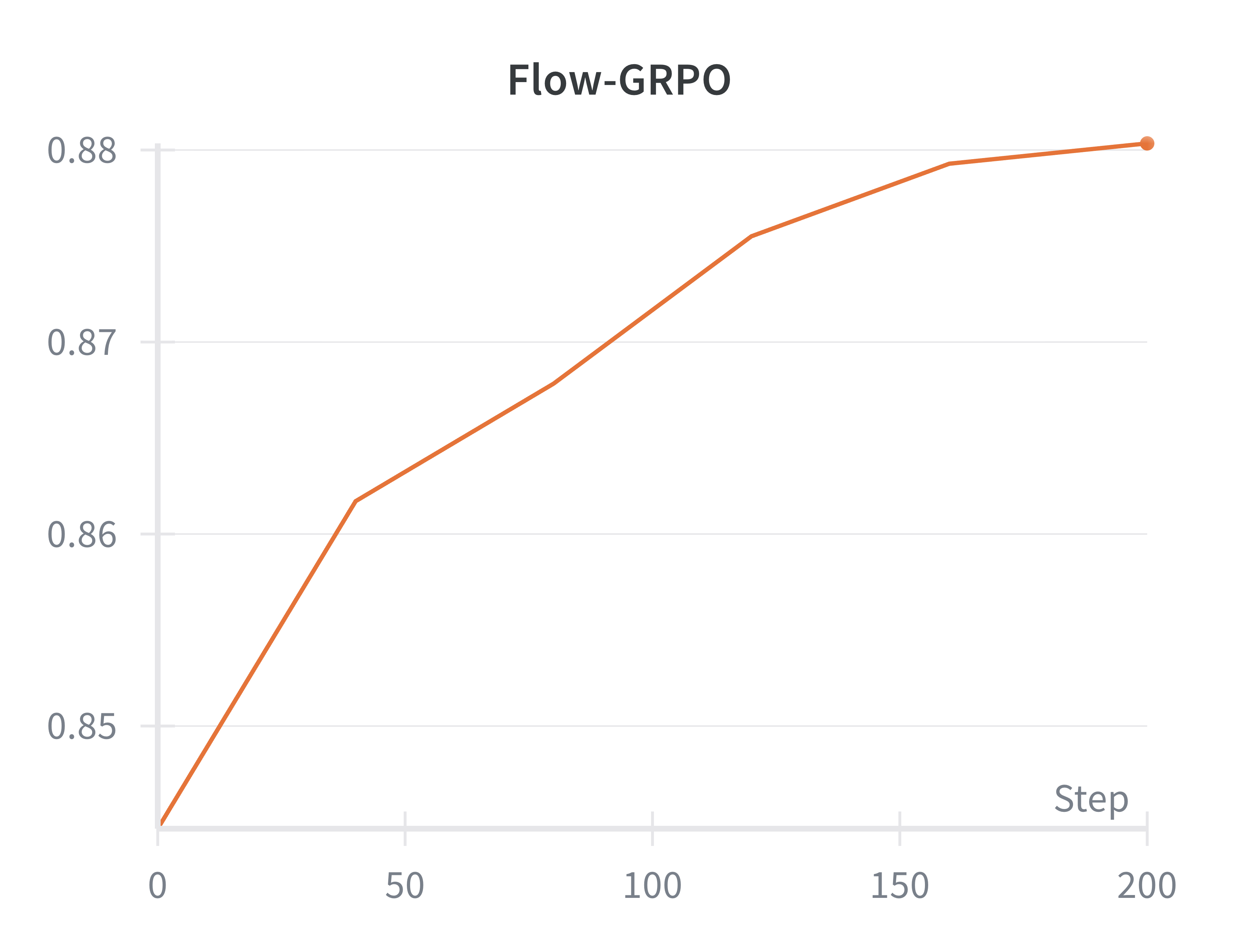}
\end{minipage}
\begin{minipage}{0.32\textwidth}
    \includegraphics[width=\linewidth]{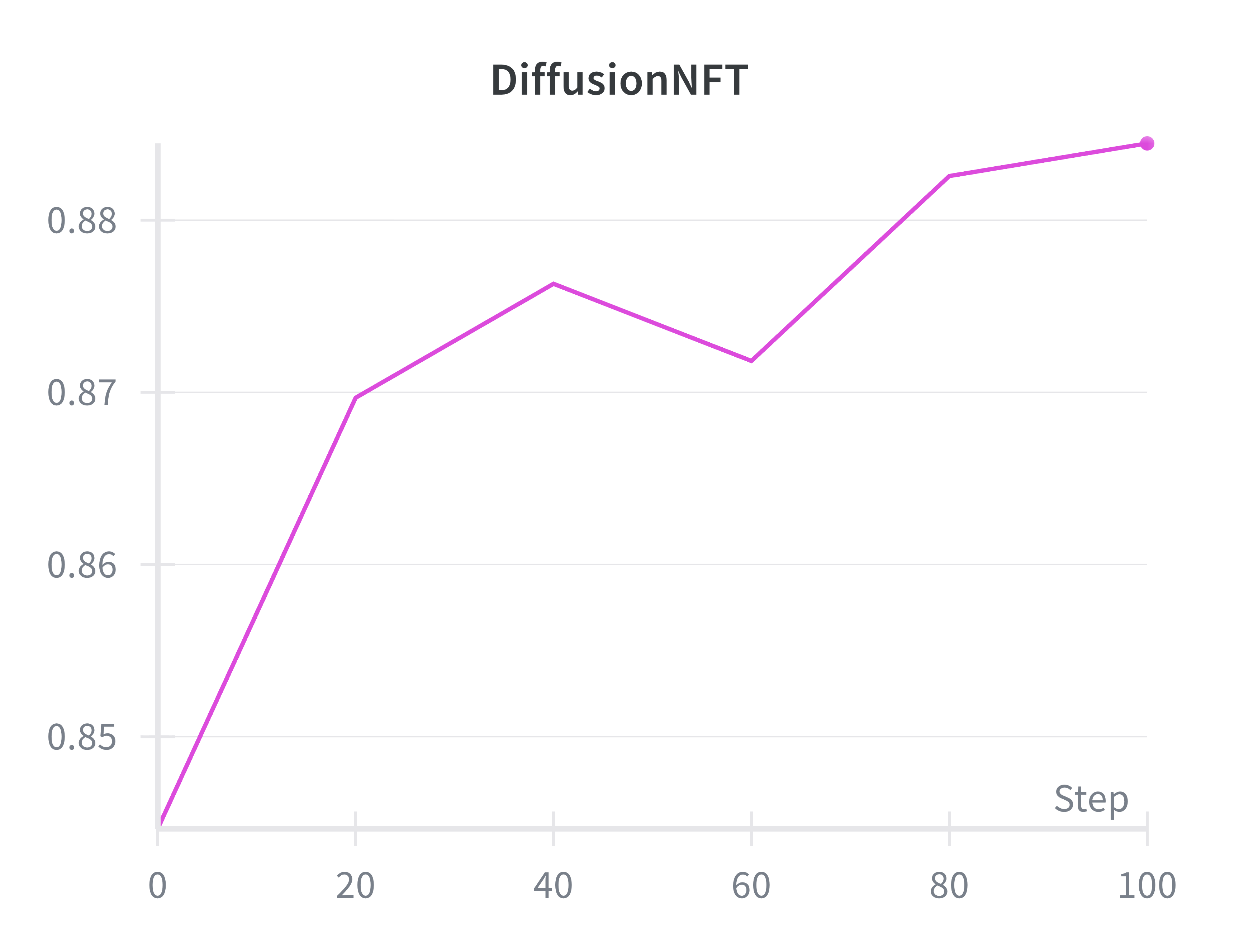}
\end{minipage}
\begin{minipage}{0.32\textwidth}
    \includegraphics[width=\linewidth]{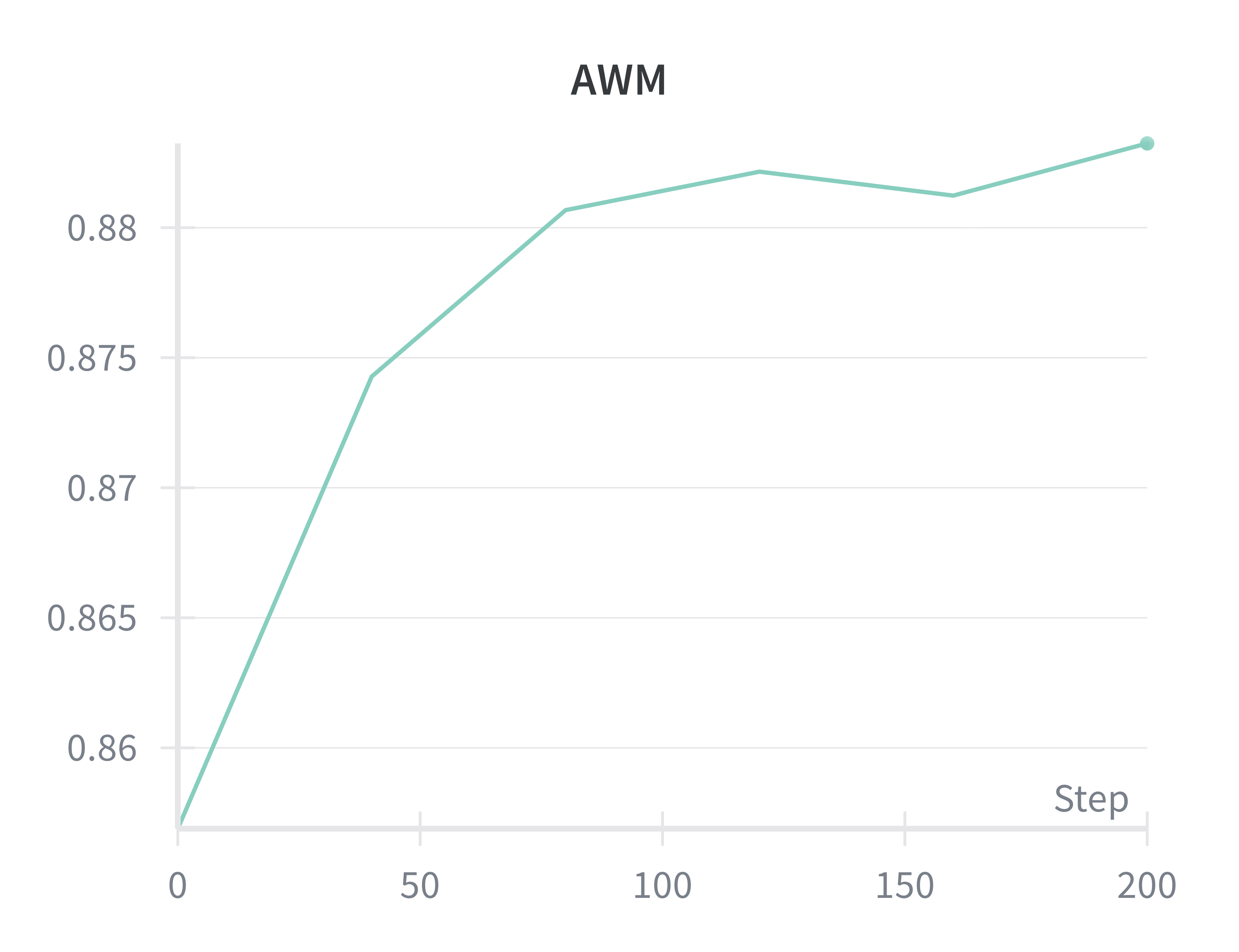}
\end{minipage}
\caption{Reproduction of reward curves for Flow-GRPO, DiffusionNFT, and AWM on Flux.1-dev with PickScore reward.}
\label{fig:reward_curves}
\end{figure}

As shown in Figure~\ref{fig:reward_curves},
Flow-Factory successfully recovers the performance gains reported in the original literature.
All three algorithms demonstrate consistent reward growth when optimized against the PickScore reward model,
achieving performance parity with their reference implementations within similar step counts.
This alignment underscores the high fidelity of our implementations,
proving that decoupling the algorithmic logic does not compromise stability or performance.

\begin{figure}[t]
\centering
\setlength{\tabcolsep}{2pt}
\renewcommand{\arraystretch}{0.8}
\begin{tabular}{c ccc}
    % Column headers (prompts)
    & \parbox{0.28\textwidth}{\centering\tiny \textit{Ultra realistic photo of a ceramic figurine of a blue fox, on a flower, professional photography, 8k, close up}}
    & \parbox{0.28\textwidth}{\centering\tiny \textit{A Peugeot 504 traveling along a dirt road in a lush green oasis in Egypt. The car is surrounded by tall date palm trees.}}
    & \parbox{0.28\textwidth}{\centering\tiny \textit{Kent Hovind DvD still from dark fantasy film 1982 conan the barbarian}} \\[4pt]
    
    % Row 1: Flux (baseline)
    \rotatebox{90}{\parbox{0.28\textwidth}{\centering\texttt{Flux}}} &
    \includegraphics[width=0.28\textwidth]{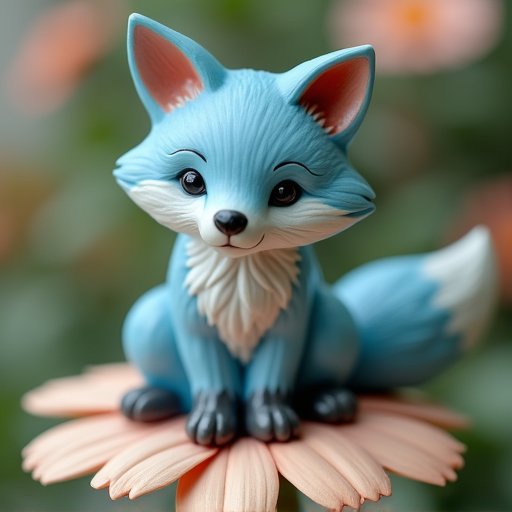} &
    \includegraphics[width=0.28\textwidth]{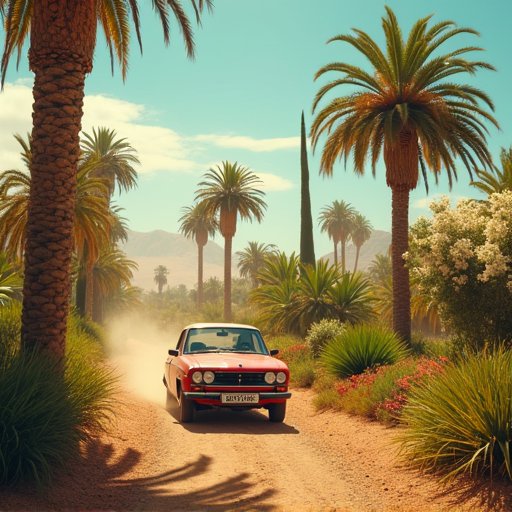} &
    \includegraphics[width=0.28\textwidth]{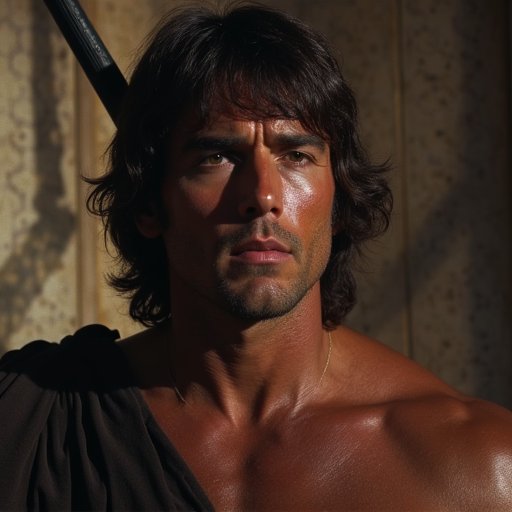} \\[2pt]
    
    % Row 2: FlowGRPO
    \rotatebox{90}{\parbox{0.28\textwidth}{\centering\texttt{FlowGRPO}}} &
    \includegraphics[width=0.28\textwidth]{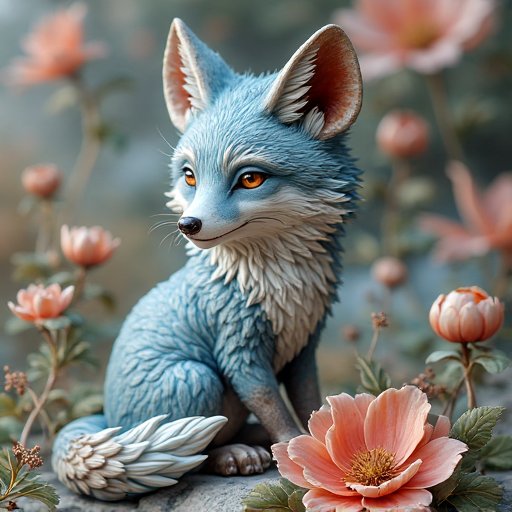} &
    \includegraphics[width=0.28\textwidth]{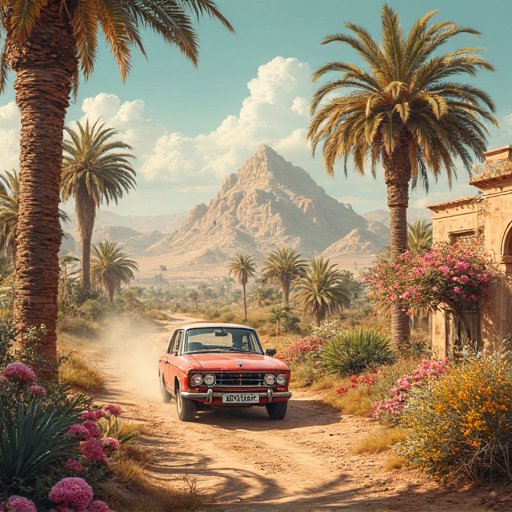} &
    \includegraphics[width=0.28\textwidth]{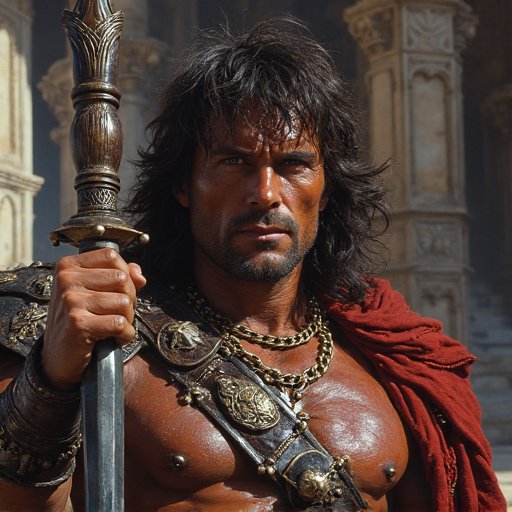} \\[2pt]
    
    % Row 3: DiffusionNFT
    \rotatebox{90}{\parbox{0.28\textwidth}{\centering\texttt{DiffusionNFT}}} &
    \includegraphics[width=0.28\textwidth]{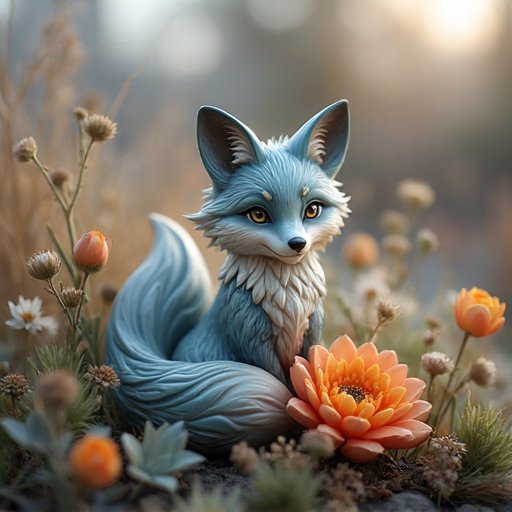} &
    \includegraphics[width=0.28\textwidth]{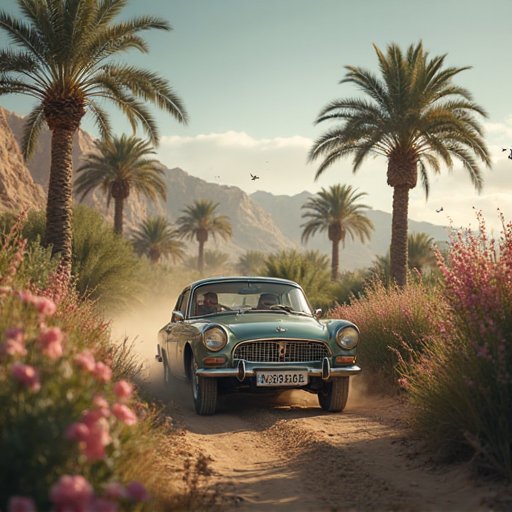} &
    \includegraphics[width=0.28\textwidth]{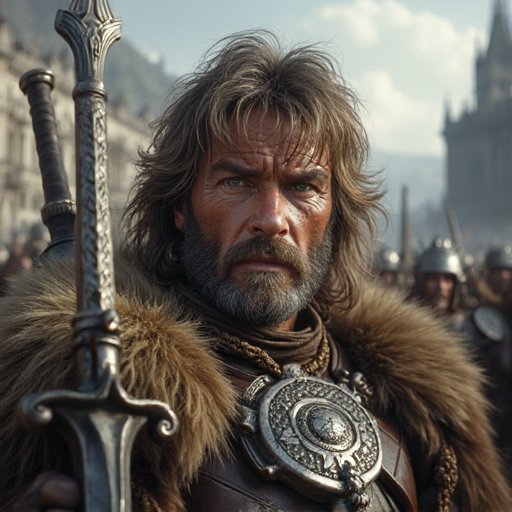} \\[2pt]
    
    % Row 4: AWM
    \rotatebox{90}{\parbox{0.28\textwidth}{\centering\texttt{AWM}}} &
    \includegraphics[width=0.28\textwidth]{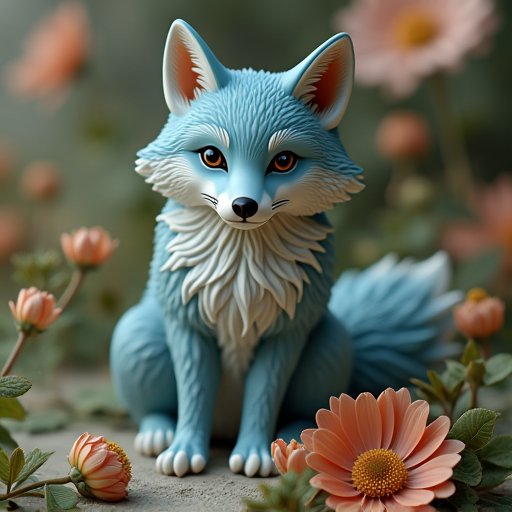} &
    \includegraphics[width=0.28\textwidth]{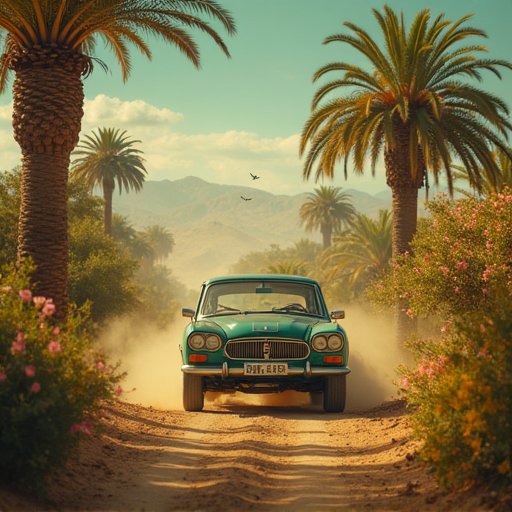} &
    \includegraphics[width=0.28\textwidth]{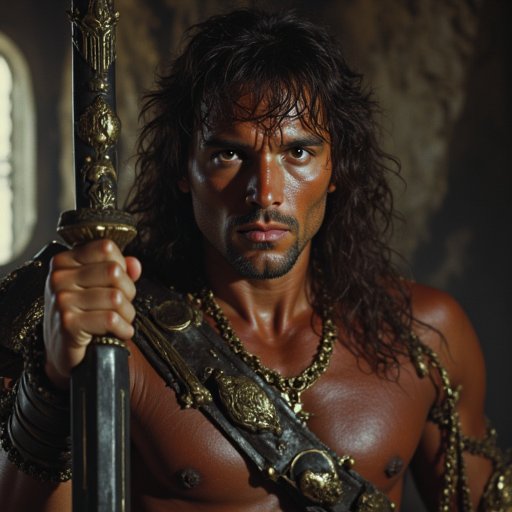} \\
\end{tabular}
\caption{Qualitative comparison of different RL algorithms on Flux.1-dev with PickScore reward. 
Each column shows generations for the same prompt across methods. 
All RL-finetuned models show improved visual quality compared to the base Flux.1-dev model.}
\label{fig:qualitative}
\end{figure}

Furthermore, we provide a qualitative comparison in Figure~\ref{fig:qualitative}.
Compared to the base FLUX.1-dev model,
all variants finetuned with the PickScore reward exhibit enhanced visual quality and stronger alignment with human aesthetic preferences.
The fact that these consistent improvements are achieved by simply switching the argument \texttt{trainer\_type} in the configuration validates Flow-Factory as a robust and extensible platform for preference alignment.

\subsection{Training Efficiency}

We evaluate the impact of preprocessing-based memory optimization on training efficiency using Flux.1-dev with PickScore reward on 8$\times$H200 GPUs (140GB each).
Each experiment runs for 100 training steps.
Table~\ref{tab:efficiency} summarizes the results.

\begin{table}[h]
\centering
\caption{Training efficiency comparison with and without preprocessing on Flux.1-dev (100 steps).}
\label{tab:efficiency}
\begin{tabular}{@{}lcc@{}}
\toprule
\textbf{Metric} & \textbf{Without Preprocessing} & \textbf{With Preprocessing} \\
\midrule
Peak GPU Memory (per device) & 61.08 GB & 53.14 GB ({\color{red}$\downarrow$13.0\%})\\
Time per Step & 144.02 s & 82.68 s ({\color{red}1.74$\times$})\\
\bottomrule
\end{tabular}
\end{table}

Preprocessing reduces GPU memory consumption by 13.0\% (7.94 GB per device) by offloading frozen text encoders components after caching their outputs.
More significantly, eliminating redundant encoding operations during training yields a 1.74$\times$ speedup, reducing per-step time from 144s to 82s.
These improvements enable larger batch sizes and faster iteration cycles, making RL fine-tuning more accessible on commodity hardware.

%% ============================================================
\section{Conclusion}
\label{sec:conclusion}
%% ============================================================

We presented Flow-Factory, a unified and extensible framework that decouples algorithms, models, and rewards for reinforcement learning on flow-matching models.
Its modular architecture enables flexible cross-combination of components through configuration alone, while preprocessing-based memory optimization and a flexible multi-reward system ensure training efficiency and algorithmic generality.
Experiments confirm that Flow-Factory faithfully reproduces state-of-the-art results with significant efficiency gains.
We hope Flow-Factory lowers the engineering barrier and accelerates community research on RL-based alignment of generative models.

\section*{Acknowledgments}
\label{sec:ack}
This work is supported by the Fundamental and Interdisciplinary Disciplines Breakthrough Plan of the Ministry of Education of China (No. JYB2025XDXM101), the National Natural Science Foundation of China (No. 62272374, No. 62192781), the Natural Science Foundation of Shaanxi Province (No.2024JC-JCQN-62), the State Key Laboratory of Communication Content Cognition under Grant No. A202502, the Key Research and Development Project in Shaanxi Province (No. 2023GXLH-024)
\textbf{and}
by A*STAR Career Development Fund <Project No. C243512010>.

\bibliography{references}
\bibliographystyle{plainnat}

\end{document}